\documentclass[10pt, letterpaper]{article}
\usepackage[paper=letterpaper, top=0.75in,bottom=0.75in,right=0.75in,left=0.75in]{geometry}

\usepackage{float}

\usepackage{cite}
\usepackage{subfig}
\usepackage[pdftex]{graphicx}

\usepackage{amsmath}
\usepackage{multirow}
\usepackage{authblk}
\begin{document}

\title{Generating Task-specific Robotic Grasps}

\author{Mark Robson}
\author{Mohan Sridharan}
\affil{Intelligent Robotics Lab\\
School of Computer Science\\
University of Birmingham, UK\\
Email: mxr880@student.bham.ac.uk,
m.sridharan@bham.ac.uk}

\maketitle

\begin{abstract}
This paper describes a method for generating robot grasps by jointly considering stability and other task and object-specific constraints. We introduce a three-level representation that is acquired for each object class from a small number of exemplars of objects, tasks, and relevant grasps. The representation encodes task-specific knowledge for each object class as a relationship between a keypoint skeleton and suitable grasp points that is preserved despite intra-class variations in scale and orientation. The learned models are queried at run time by a simple sampling-based method to guide the generation of grasps that balance task and stability constraints. We ground and evaluate our method in the context of a Franka Emika Panda robot assisting a human in picking tabletop objects for which the robot does not have prior CAD models. Experimental results demonstrate that in comparison with a baseline method that only focuses on stability, our method is able to provide suitable grasps for different tasks.

Keywords - Robot grasping, visual object representation, learning and adaptive systems.
\end{abstract}

\maketitle

\section{Introduction}
\label{sec:intro}
\begin{figure}[tb]
\centering
\subfloat[Robot set up]{\includegraphics[width=1.6in]{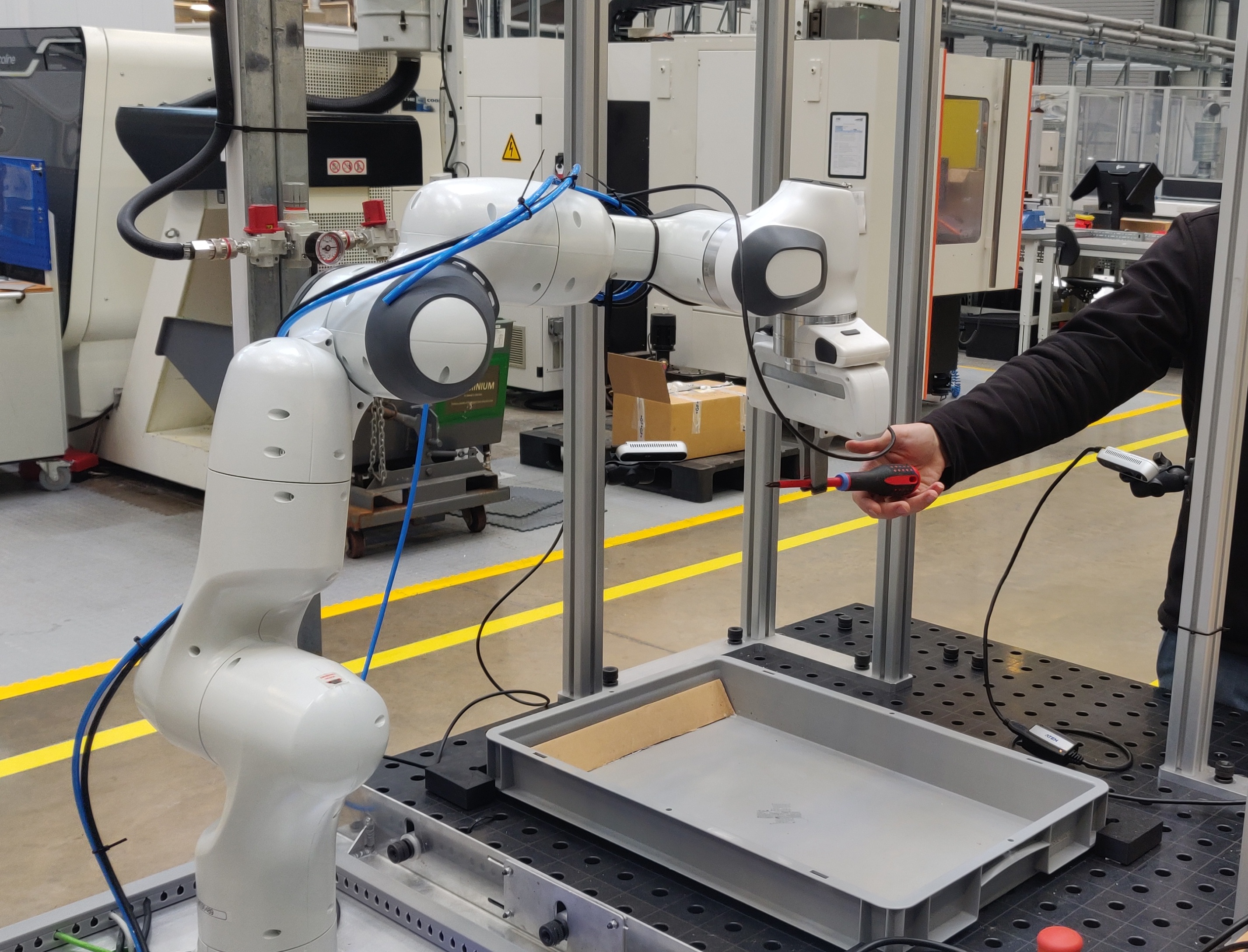}
\label{fig_root_setup}}
\subfloat[Object Set]{\includegraphics[width=1.52in]{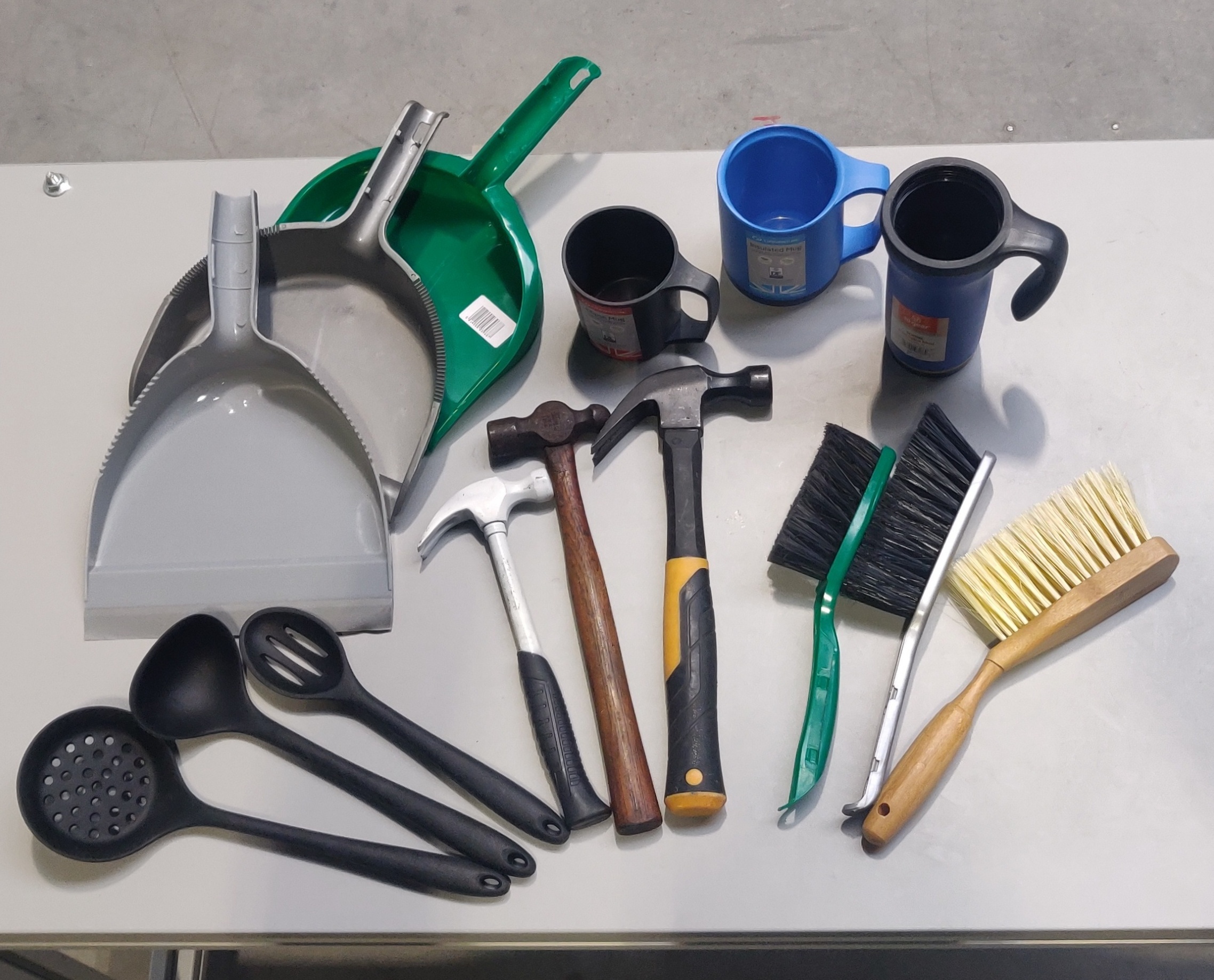}
\label{fig_objects}}
\vspace{-0.5em}
\caption{Franka manipulator robot and some objects used in the experiments. Where to grasp an object is dependent on task-specific and object-specific constraints in addition to stability.}
\label{fig_robot_objects}
\vspace{-1.25em}
\end{figure}

\vspace{-0.55em}
Consider the robot manipulator in Fig~\ref{fig_robot_objects} performing different tasks that involve interacting with relevant domain objects. Grasping an object involves contact between the fingers of the robot's hand and the object~\cite{Okamura:2000:ODM}. State of the art algorithms consider stability as a key criterion in computing suitable grasp locations on the object for the robot's fingers. However, not all stable grasps are suitable for the subsequent manipulation tasks. For example, if the task is to hand a screwdriver to a human collaborator, one would expect the robot to grasp the metallic part of the screwdriver close to the sharp tip. This is not a good location in terms of stability or when the robot has to use the tool to tighten a screw. In a similar manner, the suitable location to grasp a cup depends on whether the task is to hand it to a human or to pour its contents into a bowl.

Existing work on specifying task-specific constraints and regions for grasping objects use data-driven methods to learn from simulation trials or to define task-specific approach vectors~\cite{Kokic:ADT:2017}. There is also work on associating a semantic keypoint skeleton to an object class to identify grasp poses for specific object classes~\cite{Luo:SKP:2022,Manuelli:KPA:2019}. However, existing methods require many labelled examples and do not encode constraints that jointly consider different tasks, objects, and grasp poses. In this paper, we present an approach that seeks to address this limitation by making the following contributions:
\begin{itemize}
    \item A three-level representation of objects in terms of class membership, a semantic skeleton of keypoints, and a point cloud distribution.
    \item A method for encoding knowledge of grasps in the form of task and object-specific constraints learned from a single exemplar object and modelled to preserve the relationship between keypoints and grasp locations despite changes in scale and orientation.
    \item A simple sampling-based method that queries the learned models at run time to generate grasps which balance the task and stability constraints.
\end{itemize}

\noindent
We implement and evaluate our approach in the context of a Franka manipulator assisting humans by grasping objects on a tabletop. We show experimentally that in comparison with a baseline method that focuses on stability, our formulation also considers other task-specific criteria and provides grasps that are robust to intra-class variations in shape and appearance.

\section{Related Work}
\label{sec:relwork}
Robot grasping refers to the use of a gripper with two or more fingers to grasp an object to perform tasks of interest. State of the art methods for achieving stable grasps have transitioned in the last decade from analytic methods to data-driven methods~\cite{Bohg:DDG:2014}. Methods using deep neural networks and RGBD images have reported high success rates for grasping novel objects with a two finger parallel gripper~\cite{Morrision:CLR:2018,Song:DRGP:2022,Yang:ABRG:2021,Zeng:RPP:2017}.

Since grippers with multiple degrees of freedom pose a high-dimensional search problem, dimensionality reduction methods have been explored to identify good grasps that tend to cluster in this search space~\cite{Ciocarlie:DGE:2007, Deng:APF:2021}. Generative methods have also been developed to sample the space of grasp candidates, with some methods initialising grasp candidates based on previously successful grasps before optimising for hand configuration and finger placement~\cite{Arruda:AVD:2016,Arruda:GGS:2019,Kopicki:OSL:2015,Lu:MGT:2019}. It is difficult to compare these methods because grasp success measures are often task and domain dependent.

The choice of a good grasp is essential for the success of automated robot grasping and manipulation~\cite{Holladay:OPI:2013,Li:TOO:1988}. Considering task requirements may result in a less optimal grasp in terms of stability but it may increase the ability to manipulate the object as required by the task~\cite{Ghalamzan:TRG:2016,Zacharias:KRP:2012}. Knowledge of manipulation trajectory can also be used to impose checks on kinematic feasibility of grasps at the start and end poses, and reduce the grasp space considerably~\cite{Quispe:GFP:2016}.

Methods developed to specify task-specific constraints and regions suitable for grasping an object for a given task learn these regions from simulation trials~\cite{Berenson:CMP:2011}, from large numbers of labelled images \cite{Qian:GPDA:2020}, or as an abstract function that defines task-specific approach vectors for object classes~\cite{Kokic:ADT:2017,Song:TBR:2015}. Other work has explored the construction of a semantic keypoint skeleton to an object class~\cite{Gao:kpam2:2021,Manuelli:KPA:2019}. They use the skeleton to parameterize a single grasp pose for an object class using the keypoint skeleton for motion planning and control, but do not build a multi-level object representation, and do not explicitly model or reason about constraints that affect grasp quality. Our method, on the other hand, leverages a keypoint representation to encode task knowledge acquired from a very small set of exemplars in a model that jointly considers task- and object class-specific constraints to score grasp candidates; it is compatible with any sampling-based grasp generation method. 

\section{Problem Formulation and Approach}
\label{sec:probform-appr}
This section describes our three-level representation of objects (Section~\ref{sec:probform-appr-represent}), which guides the selection of grasps (Section~\ref{sec:probform-appr-graspcriteria}), the learning of object class- and task-specific constraints for grasp generation in addition to grasp stability (Section~\ref{sec:probform-appr-constraints}), and the sampling-based approach for the generation of grasps (Section~\ref{sec:probform-appr-graspgen}).

\subsection{Object Representation}
\label{sec:probform-appr-represent}
Our three-level object representation is based on a single exemplar for each object class. At the top level, any object is an instance of a particular class (e.g., cup or hammer). The lowest level is a point cloud representation of an object's surfaces, encoding the geometric data (e.g., shape, size) necessary for planning grasps. The intermediate level is a small set of keypoints that concisely define the object's pose while allowing for in-class variations in shape and size.  

The point cloud representation of an object is based on the probabilistic signed distance function (pSDF)~\cite{Dietrich:PMS:2016}, which allows multiple views of an object to be combined and models measurement uncertainty. A voxel grid is used to capture the signed distance to the surface from the center of each voxel. Distance data is only stored for voxels close to the measured surface using a truncation threshold. Instead of modeling distance uncertainty as a weight~\cite{Curless:VMB:1996}, surface distance is modeled as a random variable with a normal distribution \(\mathcal{N}(\overline{d}_{sdf},\,\sigma_{sdf}^{2})\). The effect of any new measurement (\(d_{sens}, \sigma_{sens}^2\)) at step $k$ is merged using a Gaussian update:
\begin{align} \label{eq:tsdfupdate}
\overline{d_k} &=\frac{\overline{d}_{k-1}.\sigma_{sens,k}^2+\overline{d}_{sens,k}.\sigma_{k-1}^2}{\sigma_{k-1}^2+\sigma_{sens,k}^2} \\ \nonumber
\sigma_{k}^2 &=\frac{\sigma_{k-1}^2.\sigma_{sens,k}^2}{\sigma_{k-1}^2+\sigma_{sens,k}^2} 
\end{align}
where the measurement variance (\(\sigma_{sens}^2\)) is computed experimentally (see Equation~\ref{eq:etheta}). We implement this encoding by revising the Open3D library~\cite{Zhou:Open3d:2018}. A uniformly-sampled point cloud, with an uncertainty measure at each point, is then extracted from the pSDF representation using the marching cubes algorithm~\cite{Lorensen:MCH:1987}. 

\begin{figure}[tb]
    \centering
    \includegraphics[width=3.3in]{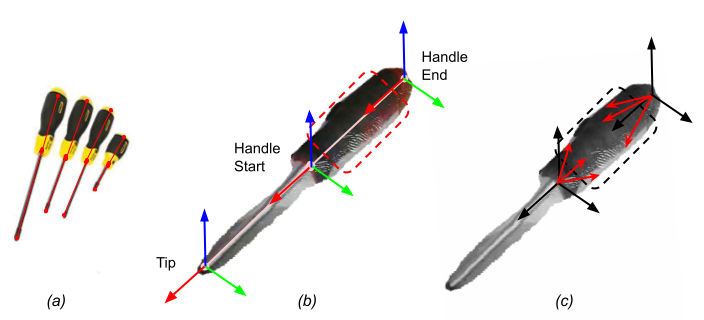}
    \vspace{-1em}
    \caption{Keypoint representation: (a) illustrates in-class variation; (b) shows coordinate systems mapped to keypoints, a domain expert associates good grasps for fastening task with points in dashed region; (c) vectors between relevant keypoints and good surface points.}
    \label{fig:keypoints}
    \vspace{-1.5em}
\end{figure}

Our intermediate representation is a lightweight set of semantic keypoints that builds on existing work~\cite{Gao:kpam2:2021,Manuelli:KPA:2019} to model large in-class variations---Figure~\ref{fig:keypoints}(b) illustrates this for screwdrivers---and provide an object pose estimate in terms of the spatial relationships between keypoints. In our implementation, 2D keypoints are detected in each colour image using a stacked hourglass network originally developed for human pose estimation~\cite{Newell:SHN:2016}. We triangulate the 3D positions of keypoints from multiple views, with links between the keypoints forming a skeleton. This skeleton is used to guide grasp generation and (as we show in Section~\ref{sec:expsetup-results-results}) to transfer knowledge for grasping across similar objects when used to perform similar tasks.
\vspace{-0.2em}
\subsection{Guiding Grasp Selection}
\label{sec:probform-appr-graspcriteria}
Our objective is to identify good grasps while considering stability, measurement uncertainty, and task-specific constraints. There is existing work on considering different criteria to design and evaluate grasps with different grippers, e.g., estimating the marginal success probability \(P(S)\) of a proposed grasp \(g\), given object state \(x\), as the sum of a weighted set of probabilistic criteria $P_i$~\cite{Chen:PFU:2018}:
\begin{align}
\label{eq:weightedprobs}
P(S|x,g) =\displaystyle\sum_{i=1}^{n} w_i.P_i, \quad \displaystyle\sum_{i=1}^{n} w_i = 1 
\end{align}
where $w_i$ are the weights. We modify the original criteria to incorporate task-specific scoring functions and focus on factors relevant to any dexterous gripper. Additional functions and factors can be added to modify grasping behaviour for specific gripper designs or tasks. 

One factor of interest is a good contact angle \(\alpha\) between gripper finger contact vector and the surface normal of the contact region on the object. We estimate the probability of good contacts for grasps with any number \(k\) of finger tips:
\begin{align} 
\label{eq:contactangleprob}
P_1 &= \left\{\begin{array}{lr}
       \displaystyle\sum_{i=1}^{k}\frac{1-\frac{2}{\theta}|\pi - \alpha_i|}{k}, & \prod_{i=1}^{k}z_i > 0\\
        0, & else
        \end{array}\right\} \\ \nonumber
z_i &=  \left\{\begin{array}{lr}
      1, & |\pi-\alpha_i|<\theta / 2\\
        0, & else
        \end{array}\right\} 
\end{align}
where \(\theta\) is the friction cone's maximum contact angle.

The next factor is based on the insight that the extent to which an object's surface is recovered from observed data contributes to the grasp success probability. Poor surface recovery makes it difficult to estimate the surface location, and a surface with a high degree of curvature or variations may not provide a good contact region. The probability of a set of surface points being good contact points is given by:
\begin{equation}  \label{eq:surfaceprob}
P_2 = \left\{\begin{array}{lr}
      \displaystyle\prod_{i=1}^k(1-\frac{c_i}{c_{max}}).(1-3u_i), & c_i<c_{max} \\
        0, & else
        \end{array}\right\}
\end{equation}
where \(u\) is a measure of the surface variation at each point based on the eigenvalue decomposition of points in the local neighbourhood; \(u=0\) if all points lie on a plane and \(u=1/3\) if points are isotropically distributed~\cite{Pauly:ESP:2002}. Also, \(c\) is an uncertainty estimate for the point from the pSDF representation with \(c_{max}\) as the maximum value; the variance is determined by experimentally set parameters. 

Finally, each task and object-specific constraint is encoded as a function that assigns a score $\in [0, 1]$ to each recovered surface point describing its likelihood of being a finger tip location for grasps that meet this constraint. The overall probability for this factor is then the product of the values of probabilistic functions for individual constraints: 
\begin{equation}\label{eq:taskobprob}
P_3 = \prod_j {T_j}
\end{equation}
In this paper, we illustrate this factor by considering one constraint per combination of task and object. Next, we describe these probabilistic functions.

\subsection{Task and Object Specific Constraints}
\label{sec:probform-appr-constraints}
While defining the functions that encode task and object-specific constraints about where to grasp an object, the objective is to model and preserve the relationship between keypoints and suitable grasp points for specific tasks across changes in factors such as scale and orientation.

Specifically, we extend the keypoint representation to include a class-specific Euclidean coordinate system at each keypoint---see Figure~\ref{fig:keypoints}(b). Consider a pair of keypoints on an object for which an exemplar grasp is available in the form of recovered grasp points on the object's surface for any particular task. The keypoints are linked by a line segment of length \(S\). The origin of the Euclidean coordinate system would be at one keypoint (the reference) with the x-axis aligned with the link.
The orientation of the y-axis and z-axis axes are defined by the plane formed by these two keypoints and one other keypoint, and the normal to the plane; if the object model has less than three keypoints, the coordinate system is based on eigenvectors of the object's point cloud. 

Given the axes at a keypoint, unit directional vectors \(\langle x,y,z\rangle = 1\) are computed to each recovered surface point of the exemplar grasp. These vectors are scaled onto a sphere of radius \(S\) and converted into spherical coordinates \(\theta, \phi \) with:
\begin{align} 
\label{eq:sphericalcoords}
S^2 = x^2 + &y^2 + z^2 \\ \nonumber
\tan\theta = y/x, \quad  
&\phi = \arccos\bigg(\frac{z}{S}\bigg)
\end{align}
where \(\theta\) $\in [-2\pi, 2\pi]$ and \(\phi\) $\in [-\pi, \pi]$. The scaled vectors are used to construct a Gaussian Mixture Model (GMM) for each keypoint using an existing software implementation of Dirichlet Process inference and Expectation Maximisation~\cite{scikitlearn}. Our representation based on spherical coordinates provides some robustness to scaling and orientation changes in new object instances of the corresponding class. At run time, point cloud points on the surface of an object are assigned a probabilistic score based on the GMM models of the ``relevant" keypoints, i.e., those that are on the nearest link (based on Euclidean distance).  
The object and task-specific probability value \(T_i\) for each surface point is the product of the probabilities from the relevant keypoints.

\subsection{Sampling and Generating Grasps}
\label{sec:probform-appr-graspgen}
We developed a sampling-based grasp generation algorithm. For any target object, RGBD images collected from four predefined poses are combined using the pSDF method described in Section~\ref{sec:probform-appr-represent} to obtain the object's point cloud, with the measurement error estimated as described in Section~\ref{sec:expsetup-results-setup}. A set of $45$ starting positions are sampled on the object's recovered surface (i.e., point cloud) with sample probability \(T_i\) (computed as described above).
Note that the baseline algorithm (for experimental evaluation) does not consider task knowledge, i.e., all points are equally likely to be sampled. 

For each sampled point, the robot virtually simulates the placement of a fingertip aligned to oppose the surface normal at the point. The gripper is positioned so that the body of the gripper is located above the object at rotations of 0$^{\circ}$ and +/-20$^{\circ}$ from the plane formed by the surface normal (at the sampled point) and the upward vector of the robot's base coordinate system. We use the Eigengrasp approach~\cite{Ciocarlie:DGE:2007} to close the gripper around the object until the other fingers contact the object's surface. The contact points of the grasp candidates which successfully close to form additional contacts with the object, and which do not collide with the table plane, are assigned scores using the weighted criteria described in Section~\ref{sec:probform-appr-graspcriteria}. Of the 135 samples, the best scoring grasp candidate in the virtual experiments is further optimised locally in three iterations with initial offsets in grasp rotation and position of 15$^{\circ}$ and 5mm, which are halved in each iteration. Only the best (optimised) grasp is executed on the robot. Note that this method is applicable to different gripper designs with different number of fingers.  

\section{Experimental Setup and Results}
\label{sec:expsetup-results}
Next, we describe the experimental setup and results.

\subsection{Experimental Setup}
\label{sec:expsetup-results-setup}
Physical experiments were conducted using a Franka Panda robot manipulator with low-cost Intel Realsense D415 cameras mounted on the end effector and as shown in Figure~\ref{fig_robot_objects}. Objects were placed on the gray tray to ensure it was within the robot's workspace and field of view. We segmented the object from the background in each image using a pretrained Detectron2 segmentation network~\cite{Wu:D2:2019}. The masked RGBD images were passed through the pSDF algorithm to obtain a point cloud. 

\paragraph{Estimating measurement noise}
The pSDF algorithm requires a model of measurement noise for each pixel in the input images. The measurement update variance \(\sigma_{sens}^2\) in Equation~\ref{eq:tsdfupdate} is estimated as the sum of two known sources of measurement error for the Realsense cameras, which we set to an X resolution \(X_{res}\) of 1280 pixels and a horizontal field of view (HFOV) of \(65\) degrees. The first source is the RMS error \(E_{drms}\) of depth measurement, which is the noise for a localised plane at a given depth \(d\) in mm~\cite{Grunnet-Jepsen:BKM:2019}: 
\begin{equation} \label{eq:edrms}
E_{drms} = \frac{0.08d^2}{55f} 
\end{equation} 
where \(f\) is the camera focal length in pixels:
\begin{equation} \label{eq:focallength}
f = \frac{0.5X_{res}}{\tan{(\frac{HFOV}{2})}}
\end{equation}
The second source of error is based on the angle between the ray from the camera and the surface, measured as \(\theta\) $\in [0, \pi]$ radians. This error \(E_\theta\) (mm) is estimated as described  in~\cite{Ahn:ANM:2019}: 
\begin{equation} 
\label{eq:etheta}
E_\theta=\frac{\theta}{(\frac{\pi}{2}-\theta)^2} 
\end{equation}

\begin{figure*}[tb]
    \centering
    \includegraphics[width=5in]{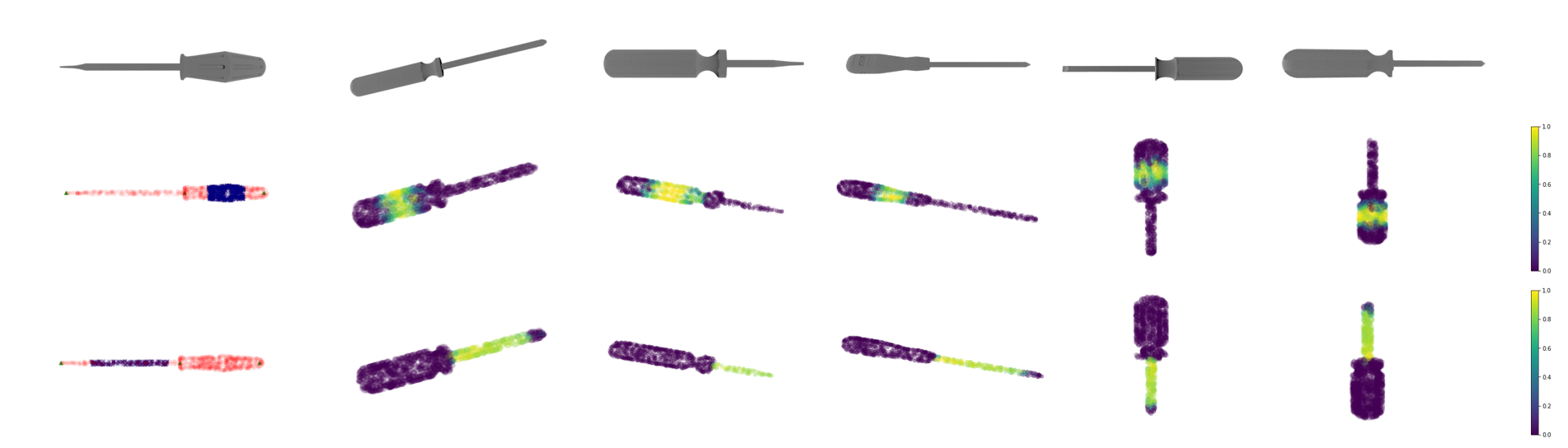}
    \vspace{-1em}
    \caption{Some instances of ``screwdriver" class. Top row shows mesh model, middle row corresponds to tool use task, and bottom row to handover task. In first column, blue points are ``good" grasp points used for training the models, green points are keypoints. Subsequent columns show score for each point on surface of point cloud using the colour scale on the right. This class has three semantic keypoints: handle start, handle end, and tip.}
    \label{fig_driver_wall}
    \vspace{-1em}
\end{figure*}

\begin{figure*}[tb]
    \centering
    \includegraphics[width=5in]{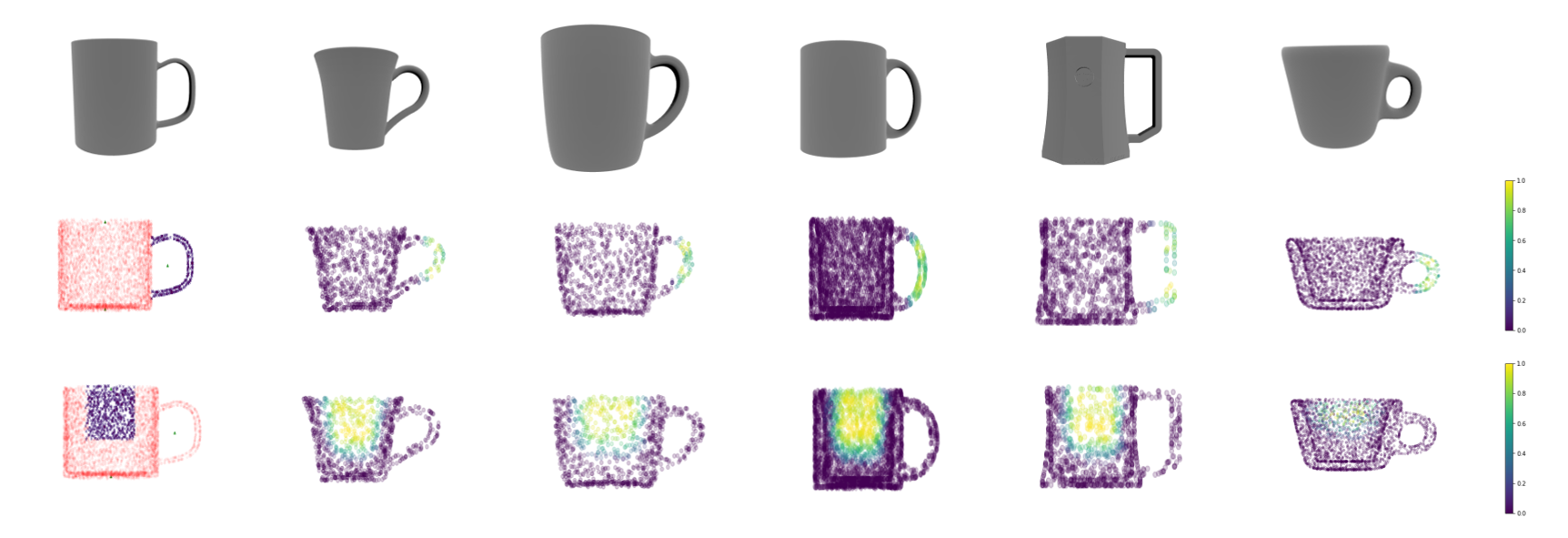}
    \vspace{-1em}
    \caption{Some instances of ``cup" class. Top row shows mesh model, middle row corresponds to pour task, bottom row to handover task. In first column blue points are ``good" grasp points used for training the models, green points are keypoints. Subsequent columns show score for each point on the surface of point cloud using the colour scale on the right. This class has three semantic keypoints: top, bottom, and handle.}
    \label{fig_cup_wall}
\end{figure*}

\paragraph{Object classes and tasks}
We considered six object classes for our physical robot experiments: cup, hammer, screwdriver, brush, dustpan, and spoon. Some instances of each class are shown in Figure~\ref{fig_robot_objects}. Object rotation was randomised to create different instances of each class; cups were, however, always placed with the opening facing upwards to enable the robot to grasp the handle. Figures~\ref{fig_driver_wall} and~\ref{fig_cup_wall} show some examples for two of the object classes.

For each class, we defined semantic keypoints (e.g., handle, top, bottom), exemplar grasps, and specific tasks. For object classes with a ``handle" region, a grasp is suitable for the ``handover" task if it leaves the handle unobstructed to allow another agent to grasp the handle when the object is presented. Each object class also supports a ``tool use" or ``pour" task. A grasp is suitable for ``pour" task if it leaves the outlet unobstructed to pour the liquid out. A grasp is suitable for ``tool use" if it leaves unobstructed the region of the object that interacts with the environment when the tool is used, e.g., head of the hammer. These grasps should also target the handle, as the handle is often designed to maximise performance when performing associated tasks. Figures~\ref{fig_driver_wall} and~\ref{fig_cup_wall} show examples of learned models for different tasks and the associated keypoints for two of the object classes. We experimentally evaluated the following hypothesis:
\begin{enumerate}
    \item[\textbf{H1}] Our method provides better grasps than baseline methods by balancing stability constraints with task/object-specific constraints;
    \item[\textbf{H2}] Our object representation supports reuse of the learned task-specific object models for other similar objects being used to perform similar tasks; and 
    \item[\textbf{H3}] Our method provides robustness to variations in size, shape, scale, and orientation within each object class.
\end{enumerate}
Hypotheses \textbf{H1} and \textbf{H2} were evaluated through robot trials, using our method to pick different objects compared with a baseline method that does not encode task-specific knowledge, i.e., it considers $P1$ (Equation~\ref{eq:contactangleprob}) and $P2$ (Equation~\ref{eq:surfaceprob}) but not $P3$ (Equation~\ref{eq:taskobprob}) in Section~\ref{sec:probform-appr-graspcriteria}; the relative weights in Equation~\ref{eq:weightedprobs} were tuned experimentally. Hypothesis \textbf{H3} was evaluated qualitatively and quantitatively by exploring the use of task models' on images depicting a range of in-class variations. In each trial, candidate grasps were generated as described in Section~\ref{sec:probform-appr-graspgen}. The robot executed the best grasp found and attempted to lift the object 10cm from the table and hold it for 10 seconds; if it succeeded, the grasp was recorded as being successful and stable. Suitability of each stable grasp to any given task was assessed visually against the given exemplars.

\renewcommand{\arraystretch}{1.2}
\begin{table*}[t]
\caption{Summary of results from 450 trials split across 5 object classes. For each class, 30 practical trials spread evenly across three example objects (see Figure~\ref{fig_objects}) were conducted for each of three task-specific models (stability, handover, tool use). In each case, the ``Stability" column indicates the proportion of successful grasps while the ``Handover" and ``Tool use" columns present the proportion of stable grasps which met the associated task criteria. Bold-faced numbers along each row indicate the best scores for the corresponding object class for each of the three tasks.}
\label{resuts_table}
\centering
\resizebox{\textwidth}{!}{\begin{tabular}{|l|l|l|l|l|l|l|l|l|l|}
\hline
\multirow{2}{*}{Object Class} & \multicolumn{3}{c|}{Baseline Model} & \multicolumn{3}{c|}{Handover Model} & \multicolumn{3}{c|}{Tool Use Model} \\ \cline{2-10} 
 & \multicolumn{1}{c|}{Stability} & \multicolumn{1}{c|}{Handover} & \multicolumn{1}{c|}{Tool Use} & \multicolumn{1}{c|}{Stability} & \multicolumn{1}{c|}{Handover} & \multicolumn{1}{c|}{Tool Use} & \multicolumn{1}{c|}{Stability} & \multicolumn{1}{c|}{Handover} & \multicolumn{1}{c|}{Tool Use} \\
\hline

Brush & 56.7\% & 52.9\% & 35.3\% & 83.3\% & \textbf{100.0\%} & 0.0\% & \textbf{90.0\%} & 0.0\% & \textbf{100.0\%} \\
Cup & \textbf{80.0\%} & 83.3\% & 12.5\% & \textbf{80.0\%} & \textbf{95.8\%} & 4.2\% & 76.7\% & 0.0\% & \textbf{82.6\%} \\
Dustpan & 86.7\% & \textbf{100.0\%} & 0.0\% & \textbf{96.7\%} & \textbf{100.0\%} & 0.0\% & \textbf{96.7\%} & 0.0\% & \textbf{100.0\%} \\
Screwdriver & \textbf{83.3\%} & 12.0\% & 72.0\% & 80.0\% & \textbf{91.7\%} & 0.0\% & \textbf{83.3\%} & 0.0\% & \textbf{96.0\%} \\
Spoon & 70.0\% & 52.4\% & 47.6\% & \textbf{90.0\%} & \textbf{88.9\%} & 11.1\% & \textbf{90.0\%} & 0.0\% & \textbf{100.0\%}\\
\hline
\end{tabular}}
\vspace{-1em}
\end{table*}
\subsection{Experimental Results}
\label{sec:expsetup-results-results}
Figures~\ref{fig_driver_wall} and~\ref{fig_cup_wall} illustrate the use of some task models from two object classes (screwdriver, cup), including the scoring of surface points of five previously unseen objects from these classes. We show just the point cloud data in these figures for ease of explanation; during run time, our algorithm processed real-world scenes containing the objects of interest.

In Figure~\ref{fig_driver_wall}, we can see that the learned models of task and object-specific constraints scale well to new object instances despite variations in the shape of the handle or length of the screwdriver shaft. The nonlinear shapes of the cups is modelled and considered when evaluating the suitability of new surface points as grasp points in Figure~\ref{fig_cup_wall}, with different sets of points being preferred for the the handover task and the pour task. These qualitative results demonstrate our method's ability to use models learned from a small set of exemplars to evaluate grasps for different tasks and guide grasping towards locations favoured by the domain expert.

\begin{figure}[b]
    \centering
    \includegraphics[width=2in]{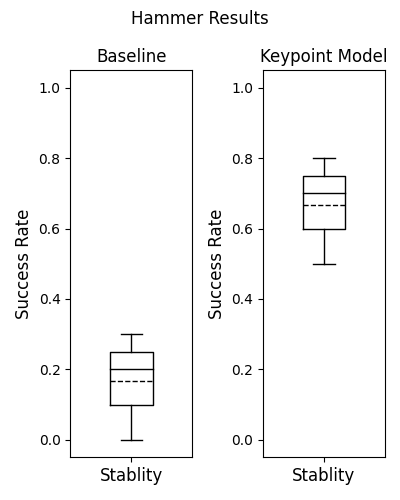}
    \vspace{-0.5em}
    \caption{Trials in the hammer class focusing on stability; trials split 20:10:10 across the three hammer objects for each of the baseline and keypoint models for a total of 80 trials.}
    \label{fig_hammer_results}
    \vspace{-1.5em}
\end{figure}

To evaluate \textbf{H1}-\textbf{H3}, we completed a total of 530 grasp trials on the physical robot platform. Table~\ref{resuts_table} summarizes results for all five classes. The results for the object class \textit{Hammer} are shown separately in Figure~\ref{fig_hammer_results} that focuses on the stability criterion to highlight some interesting results that are discussed further below.

In Table~\ref{resuts_table}, bold-faced numbers along each row indicate the best scores for the corresponding object class for each of the three tasks. For example, for the class \textit{cup}, the 'Baseline' and 'Handover' model were equally good for providing good stability, while the 'Handover' model and 'Tool use' model (i.e., the task-specific models) provide the best performance for the corresponding tasks. Our experiments showed that for some object classes the baseline algorithm is more likely to produce grasps which suit one task, e.g., handover for cups and tool use for screwdrivers. This result is expected as these tasks require grasps which place the fingers on larger, lower curvature areas of the model that better fit the stability criteria optimised by the baseline approach. Results also indicated that including task and object-specific information in the learned models for different classes steered grasps towards regions that better suit the task under consideration and produced more successful grasps.

To evaluate \textbf{H2}, we focused on the brush and dustpan object classes, which have a ``handle" configuration similar to that of the screwdriver; we modelled this configuration with a keypoint at each end. In our trials for these new object classes, we applied the same ``tool use" model which had been trained for the screwdriver. Our results show that this produced grasps with both high stability and task suitability illustrating that our method can be used to translate learned task knowledge to other similar object classes when they are used to perform similar functions.

\begin{figure}[H]
    \centering
    \includegraphics[width=5.5in]{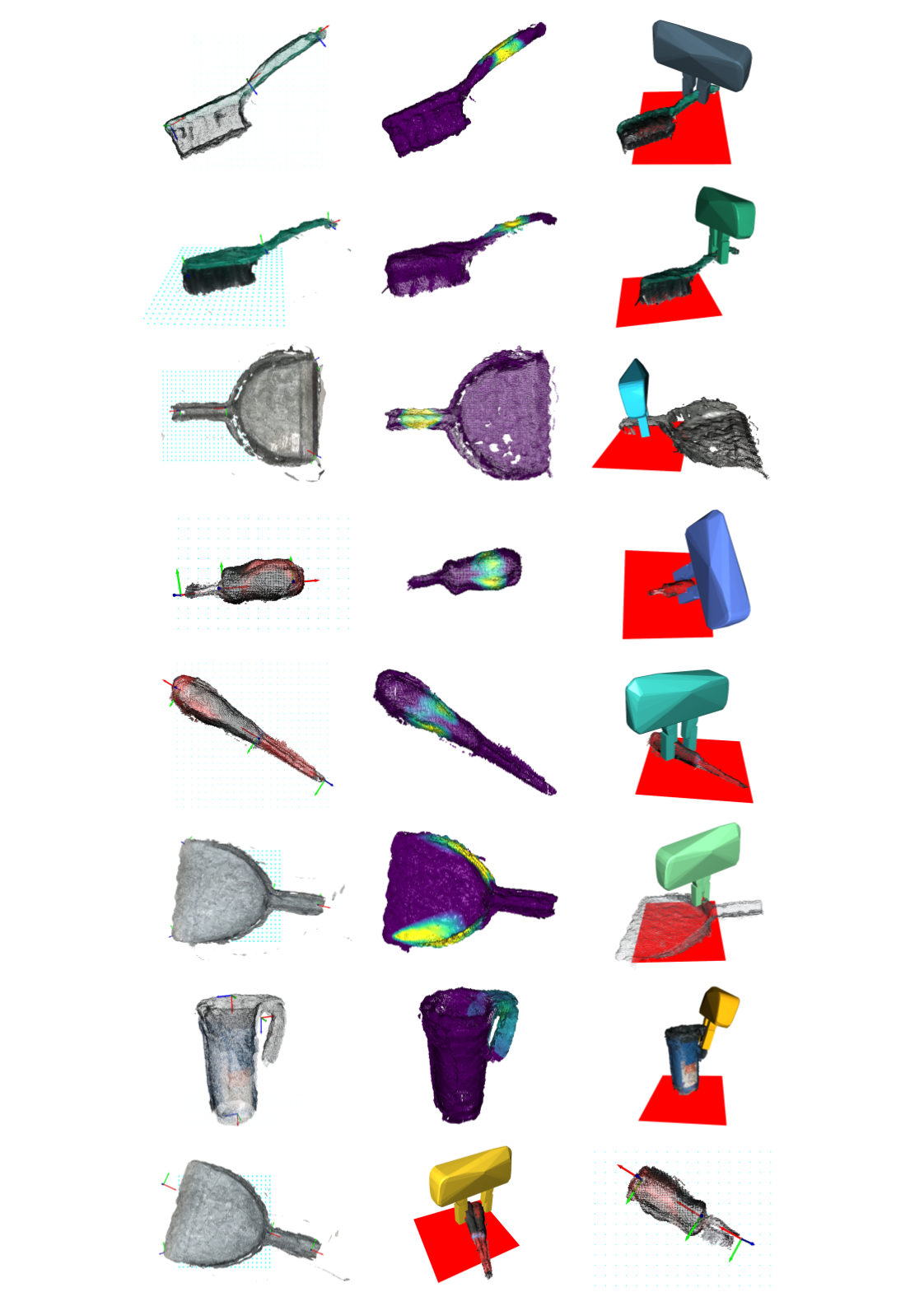}
    \caption{Illustrative examples of qualitative results. The first seven rows show use of task-specific models to guide grasping despite intra-class variations in scale and orientation: first column shows keypoints; second shows heatmap of good grasp locations (lighter colors are better); and third shows preferred grasp. Rows 1-5 show use of same task-specific model (of \textit{screwdriver} class) for the same task (tool use) across different object classes. Rows 6 and 7 show the ``handover" task model for the dustpan class and the ``pour" model for the cup class respectively. Final row shows some failure cases: first image shows incorrect keypoint detection; second shows an instance where the object moved during imaging; and third is an example of a point cloud with substantial noise, potentially leading to incorrect grasp placement.}
    \label{fig_real_pcds}
\end{figure}

The results also supported \textbf{H3}. For example, experiments in the screwdriver class showed adaption to scaling changes by stretching the learned models based on keypoint positions to handle length (size) differences. Also, the trials for the cup class illustrated the ability of our approach to handle intra-class variations in appearance and shape, with each cup having different height, diameter, thickness, and handle design. Figures~\ref{fig_driver_wall} and~\ref{fig_cup_wall} show some variations in these two classes in the point cloud representation; additional qualitative results are shown in Figure~\ref{fig_real_pcds}.

For some objects there is a small drop in stability with our approach when our task-specific models result in the robot targeting regions that the baseline algorithm is less likely to target, such as when the robot attempts to grasp the cup handle for pouring. This is expected as the handle regions typically have reduced low curvature surface area for the gripper finger tips to contact. The drop in stability is balanced by the other constraints in grasp scoring resulting in grasps with an acceptable stability which also meet the task requirements. As an additional test, we added a 6mm thick strip of foam to the handle of the cup with the thinnest handle and repeated an additional ten trials with the pour task model. With a larger surface area for grasping, the rate of successful grasps increased from 70\% to 90\% with our approach, and the proportion of these grasps that met the criteria for the pour task increased from 71\% to 88\%. Overall, these results support hypotheses \textbf{H1-H3}.

The results in Figure~\ref{fig_hammer_results} for the hammer class highlight some issues of interest. Many grasp algorithms in literature aim for the centre of objects to increase the likelihood of a stable grasp. In the case of a hammer this heuristic is flawed as the functional design of a hammer requires concentration of mass at the head end. With the baseline model and the parallel gripper, it was difficult to find a stable grasp to lift a hammer without the hammer rotating in the grasp; the robot was not able to successfully grasp, lift, and hold the hammer for the required time in many trials. The use of our learned models showed a significant improvement in stability over the baseline method because our approach learned to guide grasping away from the handle and towards the head end of the hammer shaft for stability (and handover). The functional distribution of weight and our choice of gripper make it difficult to run trials successfully for a tool use task even with our learned model; the robot often selects the correct grasp points on the handle but is unable to stably hold the object for the required time. This can be fixed by using a different gripper (e.g., more fingers that wrap around the the handle), which we will explore further in future work.

\section{Conclusions and Future Work}
\label{sec:conclusion}
We presented an approach for considering task and object-specific constraints while generating suitable grasp points on the target object's surface for robot manipulation. Unlike existing methods that focus on stability as the key criterion, our approach trades off stability with task-specific constraints on suitable grasp locations. We introduced a three-level representation for objects including class membership, point cloud data, and semantic keypoints. We also learn a model that preserves the relationship between keypoints and grasp points for specific tasks despite changes in factors such as scale and orientation of objects. Experimental evaluation on a Franka robot manipulator with a parallel gripper, with a baseline that does not consider the task-specific criteria, indicates the ability to learn from a single (or small number of) exemplar(s) provided by the designer, achieving the desired task-specific trade off and producing successful grasps of previously unseen instances of the object classes.

Our research opens up many directions for further work that address current limitations. First, we have explored the trade off of some task-specific criteria with stability; future work will include additional object classes and criteria, e.g., many task-oriented grasps align with the object's principal axes and a measure of this alignment can be use to optimise grasp locations. Second, we currently do not consider the approach vector of the hand in our grasp generation model. One extension would be to include a model of approach vectors to relevant semantic keypoints to learn task-specific grasp approach vectors. Third, the work described in this paper built an object representation based on semantic keypoints that provided some robustness to changes in scale and orientation.

Our experiments also provided promising results for the transfer of knowledge (models) learned for one object class to other classes that share similar semantic regions, e.g., handles. We will explore this transfer of knowledge in more detail in future work, using many other object classes. Finally, it would also be interesting to reduce the extent of involvement of a human expert by exploring methods that automatically segment and process image sequences (i.e., videos) to learn semantic keypoints and grasp regions for additional objects. The overall objective would be to smoothly trade off different criteria to result in safe and successful grasps for a wide range of object classes and tasks.

\bibliographystyle{IEEEtran}
\bibliography{references}

\begin{thebibliography}{10}
\providecommand{\url}[1]{#1}
\csname url@samestyle\endcsname
\providecommand{\newblock}{\relax}
\providecommand{\bibinfo}[2]{#2}
\providecommand{\BIBentrySTDinterwordspacing}{\spaceskip=0pt\relax}
\providecommand{\BIBentryALTinterwordstretchfactor}{4}
\providecommand{\BIBentryALTinterwordspacing}{\spaceskip=\fontdimen2\font plus
\BIBentryALTinterwordstretchfactor\fontdimen3\font minus
  \fontdimen4\font\relax}
\providecommand{\BIBforeignlanguage}[2]{{%
\expandafter\ifx\csname l@#1\endcsname\relax
\typeout{** WARNING: IEEEtran.bst: No hyphenation pattern has been}%
\typeout{** loaded for the language `#1'. Using the pattern for}%
\typeout{** the default language instead.}%
\else
\language=\csname l@#1\endcsname
\fi
#2}}
\providecommand{\BIBdecl}{\relax}
\BIBdecl

\bibitem{Okamura:2000:ODM}
A.~M. Okamura, N.~Smaby, and M.~R. Cutkosky, ``{An Overview of Dexterous
  Manipulation},'' in \emph{Proceedings of the 2000 IEEE International
  Conference on Robotics {\&} Automation}, no. April, 2000, pp. 255 -- 262.

\bibitem{Kokic:ADT:2017}
M.~Kokic, J.~A. Stork, J.~A. Haustein, and D.~Kragic, ``{Affordance detection
  for task-specific grasping using deep learning},'' in \emph{IEEE-RAS
  International Conference on Humanoid Robots}, no. November, 2017, pp. 91--98.

\bibitem{Luo:SKP:2022}
Z.~Luo, W.~Xue, J.~Chae, and G.~Fu, ``Skp: Semantic 3d keypoint detection for
  category-level robotic manipulation,'' \emph{IEEE Robotics and Automation
  Letters}, 2022.

\bibitem{Manuelli:KPA:2019}
\BIBentryALTinterwordspacing
L.~Manuelli, W.~Gao, P.~Florence, and R.~Tedrake, ``{kPAM: KeyPoint Affordances
  for Category-Level Robotic Manipulation},'' in \emph{International Symposium
  on Robotics Research (ISRR),}, Hanoi, Vietnam, 2019. [Online]. Available:
  \url{http://arxiv.org/abs/1903.06684}
\BIBentrySTDinterwordspacing

\bibitem{Bohg:DDG:2014}
J.~Bohg, A.~Morales, T.~Asfour, and D.~Kragic, ``{Data-driven grasp synthesis-A
  survey},'' \emph{IEEE Transactions on Robotics}, vol.~30, no.~2, pp.
  289--309, 2014.

\bibitem{Morrision:CLR:2018}
\BIBentryALTinterwordspacing
D.~Morrison, P.~Corke, and J.~Leitner, ``{Closing the Loop for Robotic
  Grasping: A Real-time, Generative Grasp Synthesis Approach},'' in
  \emph{Robotics: Science and Systems}, 2018. [Online]. Available:
  \url{http://arxiv.org/abs/1804.05172}
\BIBentrySTDinterwordspacing

\bibitem{Song:DRGP:2022}
Y.~Song, J.~Wen, D.~Liu, and C.~Yu, ``Deep robotic grasping prediction with
  hierarchical rgb-d fusion,'' \emph{International Journal of Control,
  Automation and Systems}, vol.~20, pp. 243--254, 2022.

\bibitem{Yang:ABRG:2021}
Y.~Yang, Y.~Lui, H.~Liang, X.~Lou, and C.~Choi, ``Attribute-based robotic
  grasping with one-grasp adaptation,'' in \emph{2021 IEEE International
  Conference on Robotics and Automation (ICRA)}.

\bibitem{Zeng:RPP:2017}
\BIBentryALTinterwordspacing
A.~Zeng, S.~Song, K.-T. Yu, E.~Donlon, F.~R. Hogan, M.~Bauza, D.~Ma, O.~Taylor,
  M.~Liu, E.~Romo, N.~Fazeli, F.~Alet, N.~C. Dafle, R.~Holladay, I.~Morona,
  P.~Q. Nair, D.~Green, I.~Taylor, W.~Liu, T.~Funkhouser, and A.~Rodriguez,
  ``{Robotic Pick-and-Place of Novel Objects in Clutter with Multi-Affordance
  Grasping and Cross-Domain Image Matching},'' 2017. [Online]. Available:
  \url{http://arxiv.org/abs/1710.01330}
\BIBentrySTDinterwordspacing

\bibitem{Ciocarlie:DGE:2007}
M.~Ciocarlie, C.~Goldfeder, and P.~K. Allen, ``{Dexterous Grasping via
  Eigengrasps: A Low-dimensional Approach to a High-complexity Problem},''
  \emph{Robotics Science and Systems}, 2007.

\bibitem{Deng:APF:2021}
\BIBentryALTinterwordspacing
Z.~Deng, B.~Fang, B.~He, and J.~Zhang, ``An adaptive planning framework for
  dexterous robotic grasping with grasp type detection,'' \emph{Robotics and
  Autonomous Systems}, vol. 140, 2021. [Online]. Available:
  \url{https://doi.org/10.1016/j.robot.2021.103727}
\BIBentrySTDinterwordspacing

\bibitem{Arruda:AVD:2016}
E.~Arruda, J.~Wyatt, and M.~Kopicki, ``{Active vision for dexterous grasping of
  novel objects},'' in \emph{IEEE International Conference on Intelligent
  Robots and Systems}, vol. 2016-Novem, no. August, 2016, pp. 2881--2888.

\bibitem{Arruda:GGS:2019}
\BIBentryALTinterwordspacing
E.~Arruda, C.~Zito, M.~Sridharan, M.~Kopicki, and J.~L. Wyatt, ``{Generative
  grasp synthesis from demonstration using parametric mixtures},'' in \emph{RSS
  workshop on Task-Informed Grasping}, 2019. [Online]. Available:
  \url{http://arxiv.org/abs/1906.11548}
\BIBentrySTDinterwordspacing

\bibitem{Kopicki:OSL:2015}
M.~Kopicki, R.~Detry, M.~Adjigble, R.~Stolkin, A.~Leonardis, and J.~L. Wyatt,
  ``{One-shot learning and generation of dexterous grasps for novel objects},''
  \emph{The International Journal of Robotics Research}, vol.~35, no.~8, pp.
  959--976, jul 2016.

\bibitem{Lu:MGT:2019}
Q.~Lu and T.~Hermans, ``{Modeling Grasp Type Improves Learning-Based Grasp
  Planning},'' \emph{IEEE Robotics and Automation Letters}, vol. Pre-print, pp.
  1--8, 2019.

\bibitem{Holladay:OPI:2013}
A.~Holladay, J.~Barry, L.~P. Kaelbling, and T.~Lozano-Perez, ``{Object
  placement as inverse motion planning},'' \emph{Proceedings - IEEE
  International Conference on Robotics and Automation}, pp. 3715--3721, 2013.

\bibitem{Li:TOO:1988}
Z.~Li and S.~S. Sastry, ``{Task-Oriented Optimal Grasping by Multifingered
  Robot Hands},'' \emph{IEEE Journal on Robotics and Automation}, vol.~4,
  no.~1, pp. 32--44, 1988.

\bibitem{Ghalamzan:TRG:2016}
A.~M. {Ghalamzan E.}, N.~Mavrakis, M.~Kopicki, R.~Stolkin, and A.~Leonardis,
  ``{Task-relevant grasp selection: A joint solution to planning grasps and
  manipulative motion trajectories},'' in \emph{IEEE/RSJ International
  Conference on Intelligent Robots and Systems (IROS)}.\hskip 1em plus 0.5em
  minus 0.4em\relax IEEE, oct 2016, pp. 907--914.

\bibitem{Zacharias:KRP:2012}
F.~Zacharias, \emph{{Knowledge Representations for Planning Manipulation
  Tasks}}.\hskip 1em plus 0.5em minus 0.4em\relax Springer, 2012, vol.~16.

\bibitem{Quispe:GFP:2016}
\BIBentryALTinterwordspacing
A.~H. Quispe, H.~B. Amor, H.~Christensen, and M.~Stilman, ``{Grasping for a
  Purpose: Using Task Goals for Efficient Manipulation Planning},'' 2016.
  [Online]. Available: \url{http://arxiv.org/abs/1603.04338}
\BIBentrySTDinterwordspacing

\bibitem{Berenson:CMP:2011}
\BIBentryALTinterwordspacing
D.~Berenson, ``{Constrained Manipulation Planning},'' Ph.D. dissertation,
  Carnegie Mellon Univeristy, Pittsburgh, Pennsylvania, 2011. [Online].
  Available:
  \url{https://www.ri.cmu.edu/pub{\_}files/2011/5/dmitry{\_}thesis-1.pdf}
\BIBentrySTDinterwordspacing

\bibitem{Qian:GPDA:2020}
K.~Qian, X.~Jing, Y.~Duan, B.~Zhou, F.~Fang, J.~Xia, and X.~Ma, ``Grasp pose
  detection with affordnace-based task constraint learning in single-view point
  clouds,'' \emph{Journal of Intelligent and Robotic Systems}, pp. 145--163,
  2020.

\bibitem{Song:TBR:2015}
D.~Song, C.~H. Ek, K.~Huebner, and D.~Kragic, ``{Task-Based Robot Grasp
  Planning Using Probabilistic Inference},'' \emph{IEEE Transactions on
  Robotics}, vol.~31, no.~3, pp. 546--561, 2015.

\bibitem{Gao:kpam2:2021}
\BIBentryALTinterwordspacing
W.~Gao and R.~Tedrake, ``kpam 2.0: Feedback control for category-level robotic
  manipulation,'' \emph{{IEEE} Robotics Autom. Lett.}, vol.~6, no.~2, pp.
  2962--2969, 2021. [Online]. Available:
  \url{https://doi.org/10.1109/LRA.2021.3062315}
\BIBentrySTDinterwordspacing

\bibitem{Dietrich:PMS:2016}
V.~Dietrich, D.~Chen, K.~M. Wurm, G.~V. Wichert, and P.~Ennen, ``{Probabilistic
  multi-sensor fusion based on signed distance functions},'' in
  \emph{Proceedings - IEEE International Conference on Robotics and
  Automation}, vol. 2016-June, 2016, pp. 1873--1878.

\bibitem{Curless:VMB:1996}
B.~Curless and M.~Levoy, ``{A volumetric method for building complex models
  from range images},'' \emph{Proceedings of the 23rd annual conference on
  Computer graphics and interactive techniques - SIGGRAPH '96}, pp. 303--312,
  1996.

\bibitem{Zhou:Open3d:2018}
Q.-Y. Zhou, J.~Park, and V.~Koltun, ``{Open3D}: {A} modern library for {3D}
  data processing,'' \emph{arXiv:1801.09847}, 2018.

\bibitem{Lorensen:MCH:1987}
\BIBentryALTinterwordspacing
W.~E. Lorensen and H.~E. Cline, ``Marching cubes: A high resolution 3d surface
  construction algorithm,'' \emph{SIGGRAPH Comput. Graph.}, vol.~21, no.~4, p.
  163–169, Aug. 1987. [Online]. Available:
  \url{https://doi.org/10.1145/37402.37422}
\BIBentrySTDinterwordspacing

\bibitem{Newell:SHN:2016}
A.~Newell, K.~Yang, and J.~Deng, ``{Stacked hourglass networks for human pose
  estimation},'' \emph{Lecture Notes in Computer Science (including subseries
  Lecture Notes in Artificial Intelligence and Lecture Notes in
  Bioinformatics)}, vol. 9912 LNCS, pp. 483--499, 2016.

\bibitem{Chen:PFU:2018}
D.~Chen, V.~Dietrich, Z.~Liu, and G.~von Wichert, ``{A Probabilistic Framework
  for Uncertainty-Aware High-Accuracy Precision Grasping of Unknown Objects},''
  \emph{Journal of Intelligent and Robotic Systems: Theory and Applications},
  vol.~90, no. 1-2, pp. 19--43, 2018.

\bibitem{Pauly:ESP:2002}
M.~Pauly, M.~Gross, and L.~P. Kobbelt, ``{Efficient simplification of
  point-sampled surfaces},'' \emph{Proceedings of the IEEE Visualization
  Conference}, no. Section 4, pp. 163--170, 2002.

\bibitem{scikitlearn}
Scikit-Learn, ``Gaussian mixture models,''
  \url{https://scikit-learn.org/stable/modules/mixture.html}.

\bibitem{Wu:D2:2019}
Y.~Wu, A.~Kirillov, F.~Massa, W.-Y. Lo, and R.~Girshick, ``Detectron2,''
  \url{https://github.com/facebookresearch/detectron2}, 2019.

\bibitem{Grunnet-Jepsen:BKM:2019}
A.~Grunnet-Jepsen, J.~N. Sweetser, and J.~Woodfill, ``Best-known-methods for
  tuning intel{\textregistered} realsense™ d400 depth cameras for best
  performance,'' Intel, Tech. Rep., 2019.

\bibitem{Ahn:ANM:2019}
M.~S. Ahn, H.~Chae, D.~Noh, H.~Nam, and D.~Hong, ``{Analysis and Noise Modeling
  of the Intel RealSense D435 for Mobile Robots},'' \emph{2019 16th
  International Conference on Ubiquitous Robots, UR 2019}, pp. 707--711, 2019.

\end{thebibliography}

\end{document}